\documentclass[sigconf]{acmart}
\usepackage{url}
\usepackage{booktabs}
\usepackage[linesnumbered,ruled]{algorithm2e}
\usepackage{xcolor}
\usepackage{soul}
\usepackage{enumerate}
\usepackage{enumitem}
\usepackage{xcolor}

\usepackage{amsmath,amssymb,amsfonts}
\usepackage{textcomp}
\usepackage{multirow}
\usepackage{xcolor}
\usepackage{subfigure}
\usepackage{xcolor}
\usepackage{colortbl}
\usepackage{url}

\AtBeginDocument{%
  \providecommand\BibTeX{{%
    \normalfont B\kern-0.5em{\scshape i\kern-0.25em b}\kern-0.8em\TeX}}}

\copyrightyear{2023}
\acmYear{2023}
\setcopyright{acmlicensed}\acmConference[CIKM '23]{Proceedings of the 32nd
ACM International Conference on Information and Knowledge
Management}{October 21--25, 2023}{Birmingham, United Kingdom}
\acmBooktitle{Proceedings of the 32nd ACM International Conference on
Information and Knowledge Management (CIKM '23), October 21--25, 2023,
Birmingham, United Kingdom}
\acmPrice{15.00}
\acmDOI{10.1145/3583780.3614829}
\acmISBN{979-8-4007-0124-5/23/10}

\title{Cross-city Few-Shot Traffic Forecasting via Traffic Pattern Bank}

\author{Zhanyu Liu}
\affiliation{%
  \institution{Shanghai Jiao Tong University}
  \city{Shanghai}
  \country{China}
}
\email{zhyliu00@sjtu.edu.cn}

\author{Guanjie Zheng}
\authornote{corresponding author}
\affiliation{%
  \institution{Shanghai Jiao Tong University}
  \city{Shanghai}
  \country{China}
}
\email{gjzheng@sjtu.edu.cn}

\author{Yanwei Yu}
\affiliation{%
  \institution{Ocean University of China}
  \city{Qingdao}
  \country{China}
}
\email{yuyanwei@ouc.edu.cn}



\begin{document}
\renewcommand{\shortauthors}{Zhanyu Liu, Guanjie Zheng, \& Yanwei Yu}

\begin{abstract}
Traffic forecasting is a critical service in Intelligent Transportation Systems (ITS).
Utilizing deep models to tackle this task relies heavily on data from traffic sensors or vehicle devices, while some cities might lack device support and thus have few available data.
So, it is necessary to learn from data-rich cities and transfer the knowledge to data-scarce cities in order to improve the performance of traffic forecasting.
To address this problem, we propose a cross-city few-shot traffic forecasting framework via \textbf{\underline{T}}raffic \textbf{\underline{P}}attern \textbf{\underline{B}}ank (TPB) due to that the traffic patterns are similar across cities. 
TPB utilizes a pre-trained traffic patch encoder to project raw traffic data from data-rich cities into high-dimensional space, from which a traffic pattern bank is generated through clustering.
Then, the traffic data of the data-scarce city could query the traffic pattern bank and explicit relations between them are constructed.
The metaknowledge is aggregated based on these relations and an adjacency matrix is constructed to guide a downstream spatial-temporal model in forecasting future traffic.
The frequently used meta-training framework \textit{Reptile} is adapted to find a better initial parameter for the learnable modules.
Experiments on real-world traffic datasets show that TPB outperforms existing methods and demonstrates the effectiveness of our approach in cross-city few-shot traffic forecasting.
\end{abstract}
\maketitle
\section{Introduction}

\begin{figure}[!thp]
    \centering
    \includegraphics[width=0.95\linewidth]{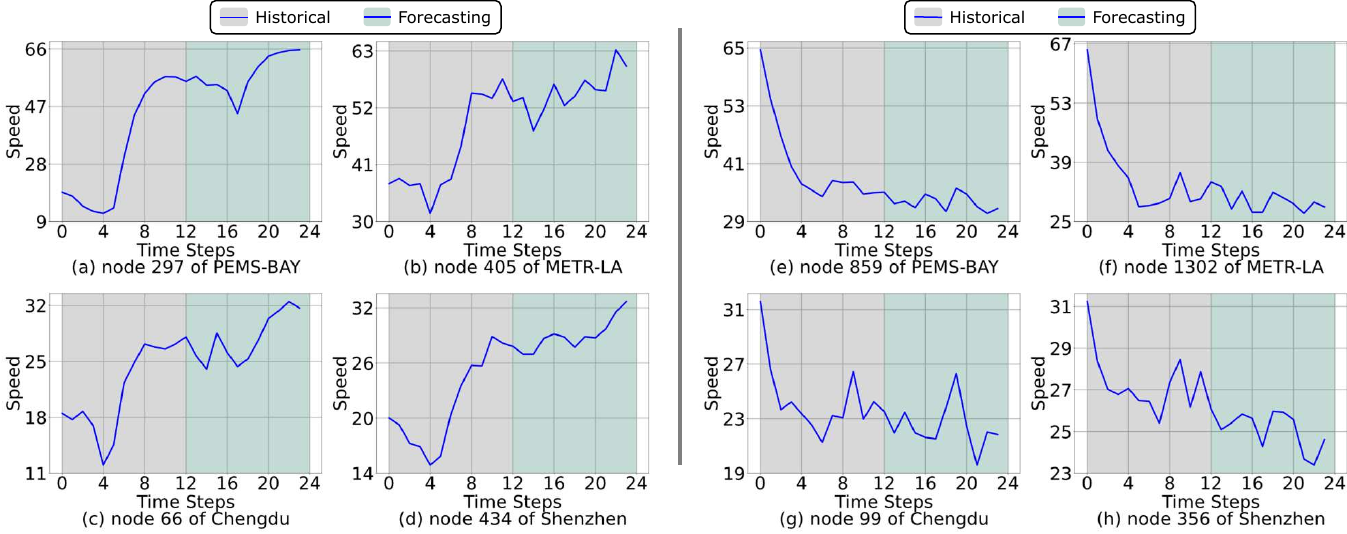}
    \caption{Examples of the observation in traffic forecasting that similar historical traffic (gray part) leads to similar future traffic (green part).  (a)$\sim$(d) The historical speed shows a rapid increase, and the future speed continues to increase. (e)$\sim$(h) The historical speed has just fallen to a low value, and the future speed fluctuates and decreases slightly.}
    \label{fig:intro}
\end{figure}

Traffic forecasting has been a critical service in Intelligent Transportation Systems (ITS).
The data used in traffic forecasting is mostly from traffic sensors~\cite{li2017diffusion} or vehicle device~\cite{Didi}.
When devices are sparse and the collected data becomes scarce, it will be challenging to train a deep model.
Consequently, learning from the data-rich cities and transferring the knowledge to the data-scarce city could be a promising solution to the cross-city few-shot traffic forecasting problem.

Currently, many studies have been working on this problem. 
RegionTrans~\cite{wang2018cross} and CrossTRes~\cite{jin2022selective} work on grid-based city data and aim to use advanced techniques to learn the region correlation between the data-rich cities and data-scarce city to transfer the knowledge better.
MetaST~\cite{yao2019learning} uses a global learnable memory to store the knowledge.
However, these methods utilize auxiliary data such as event information to transfer knowledge, which is incompatible with the city with little similar auxiliary information.
ST-MetaNet~\cite{pan2019urban} and ST-GFSL~\cite{lu2022spatio} utilize metaknowledge to generate the parameters of spatial-temporal neural networks and use the network to forecast future traffic data.
However, these methods learn the metaknowledge implicitly while ignoring the strong and explicit correlation between the traffic data of different cities and thus become less effective.

In fact, the traffic pattern is similar across cities.
Namely, when the historical traffic on roads of several cities is similar, their future traffic tends to be similar.
Fig.~\ref{fig:intro} shows some examples of the fact on four traffic datasets. 
In this figure, (a)$\sim$(d) show that a rapid increase of speed manifests the future speed will continue to increase, and (e)$\sim$(h) show a rapid decrease of speed manifests the future speed will fluctuate and decrease slightly.
If these traffic patterns could be captured, the model could explicitly relate the present traffic in the data-scarce city to the traffic patterns of data-rich cities and precisely forecast future traffic.

Inspired by this fact, we propose a novel cross-city few-shot traffic forecasting framework via \textbf{\underline{T}}raffic \textbf{\underline{P}}attern \textbf{\underline{B}}ank, which is abbreviated as \textbf{TPB}.
Specifically, we first pre-train a traffic patch encoder in data-rich cities.
The pre-trained encoder captures the general dynamics of traffic and projects the raw traffic data to a high dimensional space.
Furthermore, we input a large number of raw traffic data of data-rich cities into the pre-trained encoder to get traffic patch embeddings.
A clustering algorithm is then applied to the enormous traffic patch embeddings to eliminate redundancy and a traffic pattern bank is generated.
The traffic pattern bank contains the metaknowledge that helps forecast future traffic accurately.
Finally, the input traffic data of the data-scarce city queries the traffic pattern bank and explicit relations are constructed.
The metaknowledge is aggregated based on the relations and an adjacency matrix is constructed based on the metaknowledge to guide the downstream spatial-temporal model in forecasting future traffic.
The \textit{Reptile} meta-learning framework~\cite{nichol2018first} is utilized to find a better initial parameter of the learnable modules.
Experiments show that TPB achieves an average performance enhancement of 8.4\% and 11.3\% compared with the baselines in RMSE and MAPE respectively.

In summary, the main contributions of our work are as follows.
\begin{itemize}
    \item We investigate the cross-city few-shot traffic forecasting task and demonstrate that similar traffic patterns across cities are helpful to forecast future traffic. This explicit relation between data-rich cities and the data-scarce city is ignored by previous research.
    \item We propose a novel cross-city few-shot traffic forecasting framework TPB. The framework effectively leverages the strong and explicit correlations between traffic patterns in different cities to better forecast future traffic in the data-scarce city.
    \item We demonstrate the effectiveness of the TPB framework through extensive experiments on real-world traffic datasets. The results show that TPB is able to achieve superior performance compared to state-of-the-art methods for cross-city few-shot traffic forecasting.
\end{itemize}



\section{Related Work}
\noindent
\textbf{Traffic Forecasting}
Traffic Forecasting has been a hot research area for its important application.
Conventional traffic forecasting methods that utilize Kalman Filter~\cite{okutani1984dynamic,lippi2013short}, SVM~\cite{nikravesh2016mobile}, Bayesian Network~\cite{zhu2016short}, probalistic model~\cite{akagi2018fast}, or simulation~\cite{liang2022cblab} get good results. 
Recently the prosperity of GCN~\cite{kipf2016semi}, RNN~\cite{hochreiter1997long}, Attention~\cite{vaswani2017attention} contributes to the deep methods for modeling the spatial-temporal traffic graph thus benefiting the area of traffic forecasting.
DCRNN~\cite{li2017diffusion}, STGCN~\cite{yu2017spatio}, STDN~\cite{yao2019revisiting}, GSTNet~\cite{fang2019gstnet}, LSGCN~\cite{huang2020lsgcn}, STFGNN~\cite{li2021spatial} combine modules such as GCN, GRU, and LSTM to model the spatial-temporal relation. 
To better capture the dynamic spatial-temporal relations of the nodes of the traffic graph, methodologies such as AGCRN~\cite{bai2020adaptive}, Graph Wavenet~\cite{wu2019graph}, GMAN~\cite{zheng2020gman},  FC-GAGA~\cite{oreshkin2021fc}, D2STGNN~\cite{shao2022decoupled}, HGCN~\cite{guo2021hierarchical}, ST-WA~\cite{cirstea2022towards}, DSTAGNN~\cite{lan2022dstagnn} utilize techniques such as attention to reconstruct the adaptive adjacent matrix and fuse the temporal long term relation to make better predictions.
MTGNN~\cite{wu2020connecting} and DMSTGCN~\cite{han2021dynamic} utilize auxiliary information that helps forecast the traffic.
STGODE~\cite{fang2021spatial}, STG-NCDE~\cite{choi2022graph}, STDEN~\cite{ji2022stden} model the traffic based on ordinary differential equation(ODE).
DGCNN~\cite{diao2019dynamic}, StemGNN~\cite{cao2020spectral} view the traffic forecasting task in a graph spectral view.
PM-MemNet~\cite{lee2021learning} learns and clusters the traffic flow patterns.
STEP~\cite{shao2022pre} adapts MAE~\cite{he2022masked} and proposes a pipeline to pre-train a model.
FDTI~\cite{liu2023fdti} builds layer graphs to conduct fine-grained traffic forecasting.
However, these methods focus on single-city traffic forecasting. When facing cross-city few-shot traffic forecasting, some of their modules do not work(such as node embedding) and achieve poor results.

\noindent
\textbf{Cross-City Few-Shot Learning}
Recently Few-Shot Learning has yielded satisfying results in many areas such as computer vision~\cite{snell2017prototypical}, natural language processing~\cite{lee2022meta}, and reinforcement learning~\cite{finn2017model} when facing the problem of data scarcity.
In the area of urban computing, some methods aim to solve city data scarcity by transferring city knowledge.
Floral~\cite{wei2016transfer} transfers the knowledge of rich multimodal data and labels to the target city to conduct air quality prediction, which is mainly designed for classification problems.
RegionTrans~\cite{wang2018cross} and CrossTReS~\cite{jin2022selective} learn the region correlation between the source cities to the target city.
MetaST~\cite{yao2019learning} learns a global memory which is then queried by the target region.
STrans-GAN~\cite{zhang2022strans} utilizes the GAN-based model to generate future traffic based on traffic demand.
However, these methods focus on grid-based multimodal city region data rather than graph-based city data.
ST-MetaNet~\cite{pan2019urban} and ST-GFSL~\cite{lu2022spatio} generate the parameters of spatial-temporal neural networks according to the learned metaknowledge.
However, these methods essentially propose another way to initialize the parameter without utilizing the rich information and strong correlation between the traffic patterns of cities.

\section{Preliminary}
\textbf{Traffic Spatio-Temporal Graph:} A traffic spatio-temporal graph can be denoted as $\mathcal{G}=(\mathcal{V},\mathcal{E},\mathbf{A},\mathbf{X})$. $\mathcal{V}$ is the set of nodes and $N=|\mathcal{V}|$ is the number of nodes. $\mathcal{E}$ is the set of edges and each edge can be denoted as $e_{ij}=(v_i,v_j)$. $\mathbf{A} \in \mathbb R^{N\times N}$ is the adjacency matrix of $\mathcal{G}$, where ${a_{ij}}=1$ indicates there is an edge between $v_i$ and $v_j$. By denoting $T_{total}$ as the total number of time steps, $\mathbf{X}\in\mathbb{R}^{N\times T_{total}\times C}$ represents the node feature that contains the traffic time series data, such as traffic speed, traffic volume, and time of day.
Here, $C$ indicates the input channel and $\mathbf{X}_t\in\mathbb{R}^{N\times C}$ represents the traffic data at time step $t$.

\noindent
\textbf{Traffic Patch \& Pattern:} Traffic patch is a fixed interval traffic time series data.
For example, given traffic time series data of one day, we can split the data to get 24 one-hour traffic patches.
Formally, given traffic time series data of $v_i$ as $\mathbf{X}^i$, one traffic patch is $\mathbf{S}^i_t:= \mathbf{X}_{t:t+T_0}^i\in \mathbb{R}^{T_0\times C}$. 
In this paper, each patch contains 12 points, i.e., $T_0=12$.
Given enormous traffic patches of raw input data, the traffic patterns could be refined from them.

\noindent
\textbf{Traffic Forecasting:} Given the historical traffic data of $T$ steps, the goal of the traffic forecasting problem is to learn a function $f(\cdot)$ that forecasts the future traffic data of $T'$ steps. This task is formulated as follows.
\begin{equation}
    [\mathbf{X}^{t-T+1},\cdots,\mathbf{X}^{t}]\stackrel{f(\cdot)}{\longrightarrow}[\mathbf{X}^{t+1},\cdots,\mathbf{X}^{t+T'}]
\end{equation}

\noindent
\textbf{Cross-city Few-Shot Traffic Forecasting:} Given $P$ source cities $\mathcal{G}^{source}=\{\mathcal{G}^{source}_1,\cdots,\mathcal{G}^{source}_{P}\}$ with a large amount of traffic data and a target city $\mathcal{G}^{target}$ with only a few traffic data, the goal of cross-city few-shot traffic forecasting is to learn a model based on the available data of both $\mathcal{G}^{source}$ and $\mathcal{G}^{target}$ to conduct traffic forecasting on the future data of $\mathcal{G}^{target}$.
\section{Methodology}

\begin{figure*}[tp]
    \centering
    \includegraphics[width=0.98\linewidth]{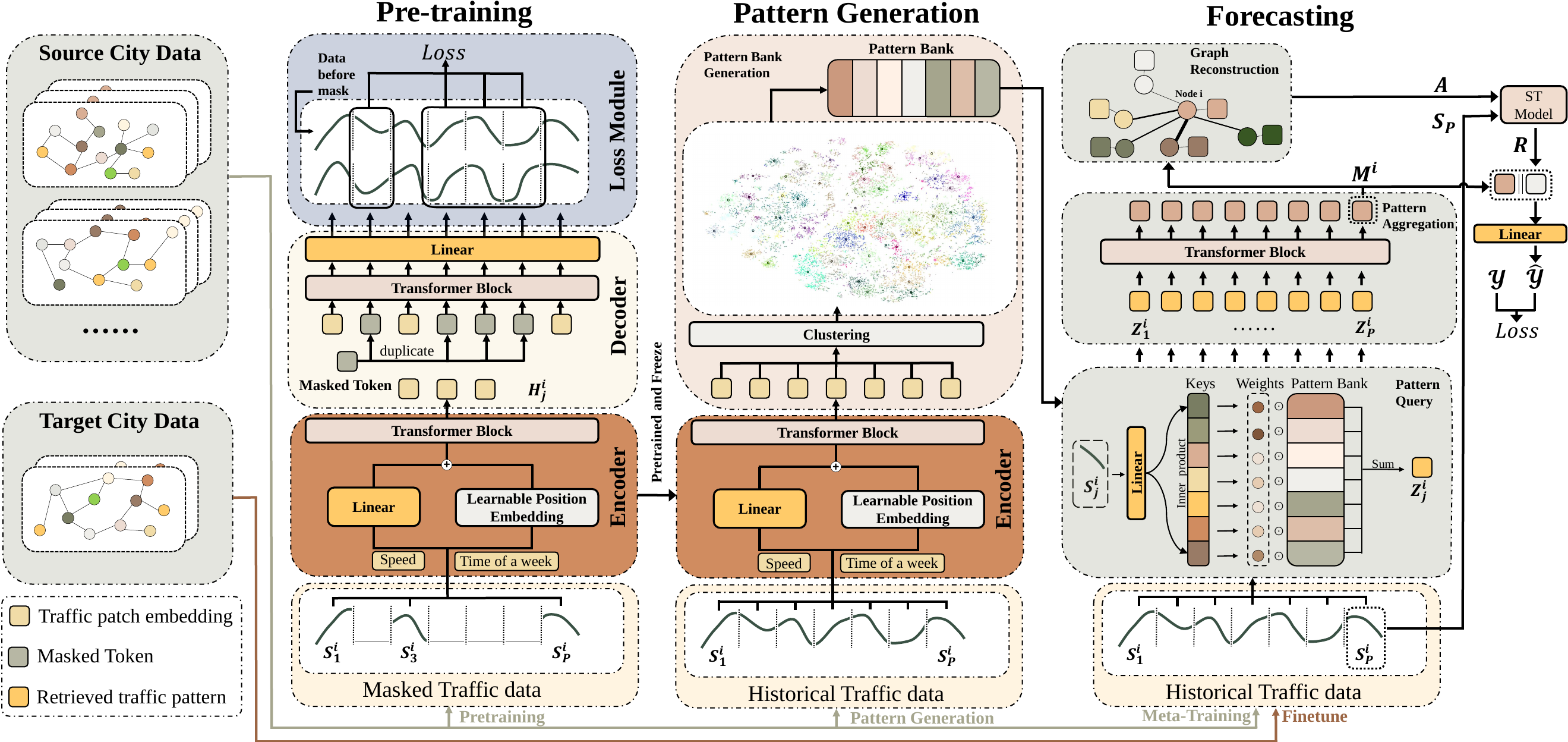}
    \caption{Diagrams of TPB. 1) In the pre-training stage, a traffic patch encoder is pre-trained by the data of source cities. 2) In the pattern generation stage, the traffic patches of source cities are put into the pre-trained encoder and the output patch embeddings are clustered to get the traffic pattern bank. 3) In the forecasting stage, the traffic pattern bank is queried by the input traffic patches and the acquired metaknowledge helps the downstream spatial-temporal model (STmodel) make accurate predictions.}
    \label{fig:TBP}
\end{figure*}

In this section, we introduce our framework \textbf{TBP} to tackle the few-shot traffic forecasting task. 
First, a traffic patch encoder is pre-trained by the data of source cities in the fashion of the Masked Autoencoder~\cite{shao2022pre,he2022masked}. 
Then, the traffic pattern bank is constructed by clustering the high-dimensional traffic patch encoder space. 
Finally, the metaknowledge is aggregated based on the traffic pattern bank and a metaknowledge-based adjacency matrix is constructed to guide a downstream spatial-temporal model to forecast future traffic.
The whole diagram is shown in Fig.~\ref{fig:TBP}

\subsection{Traffic Patch Encoder Pre-training}

\noindent{\textbf{\textit{Goal}:}}
The input raw traffic patch data of source cities contain rich information which could serve as the metaknowledge for the target city to conduct traffic few-shot forecasting.
However, the raw traffic patch is contaminated by the noise in the collecting, aggregating, splitting process, and so on.
Furthermore, the raw traffic patch lies in low-dimension space, which lacks expressiveness for downstream pattern generation. 
So, the goal of this stage is to train a traffic patch encoder that projects the input traffic patches to a higher-dimensional denoised embedding space and thus generates better traffic patch embeddings for the downstream stages.
Here, we propose a traffic patch encoder pre-training framework based on SOTA methods STEP~\cite{shao2022pre} and MAE~\cite{he2022masked}.
Other pre-training frameworks could also be adapted here.

\noindent{\textbf{Masked Input:}}
The input is $P$ continuous traffic patches of $v_i$, denoted as $\mathbf{S}^i=\{\mathbf{S}_1^i,\cdots,\mathbf{S}_P^i\}$, where the shape of each patch is $\mathbf{S}^i_j\in \mathbb{R}^{T_0\times C}$.
Here, $C$ is the input channel, $T_0$ is set to 12, and $P$ is set to 24.
Note that the interval between data points is 5 minutes so the actual length of each patch is one hour and the actual length of historical patches is one day.
We randomly set a subset of the traffic patches and mask them.
The selected patches are not required to be continuous.
We set the mask ratio to 75\% according to the original papers~\cite{shao2022pre,he2022masked}. 
Such a high mask ratio makes the pre-training task difficult so the pre-trained encoder is expected to be more robust and potentially more effective when transferred to other cities.

\noindent{\textbf{Encoder:}}
The encoder should be able to aggregate the information on the remaining parts of the traffic patches. 
The models based on RNN are not suitable here since most of the patches are masked.
Here, we adopt transformer~\cite{vaswani2017attention} as the encoder of the masked traffic patches.
Before being put into the transformer layers, each unmasked raw patch $\mathbf{S}^i_j$ is fed into a linear layer.
Then the learnable positional embedding based on the time of the week is added.
There is a total of 168 learnable positional embeddings since there are $24\times7$ hours in a week.
After adding the positional embedding, the unmasked patches are put into the transformer encoder and the corresponding output embeddings are generated.
Formally, we can write as follows.
\begin{equation}
    \mathbf{H}^i_j =  TS_E(Concat\{(\mathbf{W}_{enc}\mathbf{S}^i_j+\mathbf{b}_{enc})+\text{PE}(\mathbf{S}^i_j)|\mathsf{M}^i_j = 0\})
\end{equation}
Here, PE() indicates learnable positional embedding and $TS_E$() indicates the encoder transformer layers.
$\mathsf{M}^i_j=0$ indicates $\mathbf{S}^i_j$ is not masked and vice versa. 
$\mathbf{W}_{enc}\in\mathbb{R}^{d \times {T_0C}}$ and $\mathbf{b}_{enc}\in\mathbb{R}^d$ are learnable parameters.

\noindent{\textbf{Decoder:}}
The decoder takes in the unmasked patch embeddings to reconstruct the whole time series back to its original numerical values.
First, a learnable masked token $\mathbf{H}_{mt}\in\mathbb{R}^{d}$ is duplicated and fills the masked positions.
After filling the masked positions, the number of patches aligns with the original input.
Then, the filled patch series is put into the transformer layer to aggregate the patch series information.
Finally, the output embedding of the transformer layer is fed into a linear layer to reconstruct the original time series, noted as $\hat{\mathbf{S}}^i_j$.
Formally, the decoder process could be represented as follows.
\begin{equation}
    \hat{\mathbf{S}}^i_j = \mathbf{W}_{dec}\cdot {TS_D(Concat\{\mathbf{H}^i_j,\mathbf{H}_{mt}|\mathsf{M}^i_j=0\})} +\mathbf{b}_{dec}
\end{equation}
Here, $TS_D$() indicates the decoder transformer layer, and $\mathbf{W}_{dec}\in\mathbb{R}^{T_0C \times d }$ and $\mathbf{b}_{dec}\in\mathbb{R}^{T_0C}$ are learnable parameters.

\noindent{\textbf{Loss Module:}}
The task of the pre-training is to reconstruct the masked traffic patches while the unmasked patch is the input of the model.
So, the final loss only contains the reconstruction Mean Square Error (MSE) of masked traffic patches.
Formally, the loss of the pre-training stage is as follows.
\begin{equation}
    \mathcal{L}_{p}=\sum_{i=1}^N\sum_{j=1}^P \mathsf{M}^i_j(\mathbf{S}^i_{j} - \hat{\mathbf{S}}^i_{j})^2
\end{equation}
In summary, the pre-training stage reconstructs the mask traffic patch. 
During the pre-training, the encoder is able to learn useful features about the traffic patch and project the traffic patches to a higher-dimensional denoised embedding space, which benefits the process of pattern generation.


\subsection{Traffic Pattern Generation}
In this part, we propose a framework that generates the representative traffic patch embeddings, i.e., traffic patterns. The framework is based on the patch encoder pre-trained in the previous stage. The algorithm is shown in Alg.~\ref{alg:tpg}.

\begin{algorithm}[t]
    \KwIn{Pre-trained traffic patch encoder $TS_E$, Historical traffic patch of source cities $\mathbf{S}^i_j$, number of clusters $K$.}
    \KwOut{Traffic pattern bank $\mathbf{B}\in\mathbb{R}^{K\times d}$}

    \For{$i\leftarrow\text{range($0,N,1$)}$}{
        \For{$j\leftarrow\text{range($0,T_{total},P$)}$}{
            $\mathbf{H}^i_j,\cdots,\mathbf{H}^i_{j+P} = TS_E(\mathbf{S}^i_j,\cdots,\mathbf{S}^i_{j+P})$\;
        }
    }
    Collect $\{\mathbf{H}^i_j|i=1,...,N; j=0,...,T_{total}\}$\;    
    $o_{1:K} = KMeans(\{\mathbf{H}^i_j\})$\;
    $\mathbf{B}\leftarrow o_{1:K}$\;
    \Return $\mathbf{B}$\;

\caption{Traffic Pattern Generation}
\label{alg:tpg}
\end{algorithm}

\noindent{\textbf{\textit{Goal}:}}
The enormous traffic patches in the source cities contain rich information, which could serve as the metaknowledge for the few-shot traffic forecasting on the target city.
However, directly using the enormous traffic patches to make predictions may not be practical since the number of patches is too massive.
So, we encode the traffic patches and use clustering techniques to identify representative traffic patches to construct a small traffic pattern bank.
The small traffic pattern bank not only reduces the computational burden of the forecasting model but also serves as a compact representation of the traffic patterns in the source cities.

\noindent{\textbf{Framework:}}
We directly utilize the traffic patch encoder $TS_E$() that is pre-trained in the previous stage. 
The enormous traffic patches in the source cities $\mathbf{S}^i_j$ are put into the patch encoder to generate traffic patch embeddings $\mathbf{H}^i_j$. 
\begin{equation}
    \label{eq:Enc}
    \mathbf{H}^i_0,\mathbf{H}^i_1,\cdots,\mathbf{H}^i_P = TS_E(\mathbf{S}^i_0,\mathbf{S}^i_1,\cdots,\mathbf{S}^i_P)
\end{equation}
The pre-trained patch encoder could project the traffic patches into high-dimensional embedding space, and traffic patches with similar traffic semantics are projected to near space.

After the traffic patches are projected into the embedding space, we can then use clustering techniques to group similar traffic patches together.
Several different approaches can be taken when it comes to clustering the traffic patches, including unsupervised techniques like k-means and density-based clustering.
Here, to simplify the framework and get robust traffic patterns, we adopt \textit{k-means} as the clustering algorithm and \textit{cosine} as the distance function.
\begin{equation}
    \label{eq:KMeans}
    o_{1:K} = KMeans(Concat\{\mathbf{H}^i_j|i=1,...,N; j=0,...,T_{total}\})
\end{equation}
Here, $o_{1:K}\in\mathbb{R}^{K\times d}$ are the centroids of the clusters.

Once the traffic patches have been grouped into clusters, we can then select a representative set of traffic patterns from each cluster to construct the traffic pattern bank. 
This can be done in a variety of ways, such as selecting the mean embedding in each cluster, selecting the patch that is nearest to the centroid, or using some other metric to determine the most representative traffic pattern.
In the few-shot traffic forecasting setting, the traffic patches of the target city are unknown and some distribution bias could exist.
So, the centroid represents the general characteristics of the traffic patterns in the cluster rather than being specific to any particular traffic patch. 
We choose the centroids of each cluster $o_{1:K}$ as the traffic pattern bank $\mathbf{B}$ to guarantee robustness and transferability.
The traffic pattern bank will offer the metaknowledge in the few-shot traffic forecasting stage.

\subsection{Forecasting Stage}

In this part, we propose a framework that utilizes the metaknowledge from the traffic pattern bank learned in the data of source cities to conduct few-shot traffic forecasting.
The core idea is to query the traffic pattern bank with historical traffic patches and aggregate the retrieved patterns as the auxiliary metaknowledge.
The metaknowledge is used to reconstruct graph structure and guide the downstream spatial-temporal model to forecast future traffic.

\noindent{\textbf{\textit{Goal}:}}
The goal of this framework is to use the metaknowledge from the traffic pattern bank to improve the accuracy of few-shot traffic forecasting in the target city.

\noindent{\textbf{Pattern Query:}}
The traffic pattern bank consists of two components: \textit{Key} and Pattern Bank $\mathbf{B}$.
The \textit{Key} is a learnable embedding matrix with shape $\mathbb{R}^{K\times d_q}$, where $K$ is the number of patterns and $d_q$ is the embedding shape of \textit{Key}.
The Pattern Bank $\mathbf{B}\in\mathbb{R}^{K\times d}$ is a fixed embedding matrix learned from the previous clustering stage.

\begin{algorithm}[t]
    \KwIn{Forecasting model $F_\theta(\cdot)$, Source city data $G_{source}$, Target city data $G_{target}$}
    \KwOut{Trained Model Parameter $\theta$}
    \tcc{Meta-training the $\theta$ with $G_{source}$}
    Random initialize $\theta$\;
    \For{$e\leftarrow range(0,meta\_epochs,1)$}{
        \tcp{sample support and query tasks $\tau_{spt}$ and $\tau_{qry}$}
        $\tau_{spt}, \tau_{qry}\longleftarrow SampleTask(G_{source})$\;
        \tcp{get historical data $\mathbf{S}$ and label data $\mathcal{Y}$}
        $\mathbf{S}_{spt}, \mathcal{Y}_{spt}\longleftarrow \tau_{spt}$\;  
        $\mathbf{S}_{qry}, \mathcal{Y}_{qry}\longleftarrow \tau_{qry}$\;
        $\theta_{\tau}\longleftarrow\theta$\;
        \tcp{compute the gradient of each step}
        \For{$i\leftarrow range(0,update\_step,1)$}{
            $\hat{\mathcal{Y}}^{\theta_\tau}_{spt}\longleftarrow F_{\theta_\tau}(S_{spt})$\;
            compute $\triangledown_{\theta_\tau}\mathcal{L}_i(\hat{\mathcal{Y}}^{\theta_\tau}_{spt}, \mathcal{Y}_{spt})$ by Eq~\eqref{eq:loss}\;
            ${\theta_\tau}\longleftarrow {\theta_\tau} - \alpha\triangledown_{\theta_\tau}\mathcal{L}_i(\hat{\mathcal{Y}}^{\theta_\tau}_{spt}, \mathcal{Y}_{spt})$\;
            $\hat{\mathcal{Y}}^{\theta_\tau}_{qry}\longleftarrow F_{\theta_\tau}(S_{qry})$\;
            compute \& store $\triangledown_{\theta_\tau}\mathcal{L}_i(\hat{\mathcal{Y}}^{\theta_\tau}_{qry}, \mathcal{Y}_{qry})$\;
        }
        \tcp{use the gradients to update $\theta$}
        \For{$i\leftarrow range(0,update\_step,1)$}{
            $\theta\longleftarrow\theta - \frac{\beta}{update\_step}\triangledown_{\theta_\tau}\mathcal{L}_i(\hat{\mathcal{Y}}^{\theta_\tau}_{qry}, \mathcal{Y}_{qry})$
        }
    }  
    \tcc{Finetune the $\theta$ with $G_{target}$}
    \For{$e\leftarrow range(0,train\_epochs,1)$}{
        \For{Iterate All Batches}{
            $\mathbf{S}, \mathcal{Y}\longleftarrow SampleBatch(G_{target})$\;
            $\hat{\mathcal{Y}}^{\theta}\longleftarrow F_{\theta}(\mathbf{S})$\;
            compute $\triangledown_{\theta}\mathcal{L}(\hat{\mathcal{Y}}^{\theta}, \mathcal{Y})$ by Eq~\eqref{eq:loss}\;
            use $\triangledown_{\theta}\mathcal{L}(\hat{\mathcal{Y}}^{\theta}, \mathcal{Y})$ to update $\theta$ with Adam optimizer\;
        }
    }
    \Return $\theta$
    
\caption{Meta-training and Fine-tuning Process}
\label{alg:meta}
\end{algorithm}

\noindent Suppose the $P$ continuous traffic patches of $v_i$ are denoted as $\mathbf{S}^i=\{\mathbf{S}_1^i,\cdots,\mathbf{S}_P^i\}$.
For each patch $\mathbf{S}^i_j$, it will be used as the query to the traffic pattern bank.
To do this, we first compute the dot product scores $\omega$ between the linear transformed query patch and each key in the \textit{Key} matrix.
\begin{equation}
    \omega^k = dot(\mathbf{W}_q\mathbf{S}^i_j+\mathbf{b}_q,Key^k)
\end{equation}
Here, $\mathbf{W}_q\in\mathbb{R}^{d_q\times d}$ and $\mathbf{b}_q\in\mathbb{R}^{d_q}$ represents the linear layer. $Key^k\in\mathbb{R}^{d_q}$ is the \textit{k-th} key of the traffic pattern and $\omega^k\in\mathbb{R}$ is the score. Then, the retrieved traffic pattern $\mathbf{Z}^i_j\in\mathbb{R}^d$ could be represented as a weighted sum of traffic pattern $\mathbf{B}^k\in\mathbb{R}^d$ as follows.
\begin{equation}
    \mathbf{Z}^i_j=\sum_{k=1}^K \omega^k\cdot \mathbf{B}^k
\end{equation}
This process allows the model to retrieve the most relevant traffic patterns for a given query patch, which can then be used for further metaknowledge generation.

\noindent{\textbf{Pattern Aggregation:}}
Once the retrieved traffic pattern $\mathbf{Z}^i_j$ has been identified for each patch $\mathbf{S}^i_j$, the metaknowledge $\mathbf{M}^i$ of the historical traffic data is extracted from the retrieved traffic pattern series.
We use transformer layers to achieve this goal.
Formally, by denoting the retrieved traffic pattern series as $\{\mathbf{Z}^i_1,...,\mathbf{Z}^i_P\}$, the metaknowledge $\mathbf{M}^i\in\mathbb{R}^k$ is represented as the last hidden states of the output of the transformer layer, which is shown as follows.
\begin{equation}
    \mathbf{M}^i=Last(TS(\{\mathbf{Z}^i_1,...,\mathbf{Z}^i_P\}))
\end{equation}
Here, \textit{TS}() is the transformer layer and \textit{Last}() takes the last hidden states of the output of the transformer.

\noindent{\textbf{Graph Reconstruction:}}
To better describe the correlations between the nodes in the target city, we reconstruct an adaptive adjacency matrix based on the metaknowledge $\mathbf{M}^i$ of each node.
Here we did not use the original prior-knowledge-based graph since the adaptive graph is more effective in the traffic forecasting scenarios~\cite{liu2023really}.
The reconstructed adjacency matrix $\mathbf{A}\in\mathbb{R}^{N\times N}$ is a weighted representation of the connectivity between nodes in the graph, where $N$ is the number of nodes in the graph.
To reconstruct the adjacency matrix, we first linearly project the metaknowledge $\mathbf{M}^i$ to query space and key space.
\begin{equation}
  \begin{aligned}
    \mathbf{Q}^i=\mathbf{W}_Q\mathbf{M}^i+\mathbf{b}_Q\\
    \mathbf{K}^i=\mathbf{W}_K\mathbf{M}^i+\mathbf{b}_K
    \end{aligned}
\end{equation}
Here, $\mathbf{W}_Q,\mathbf{W}_K\in\mathbb{R}^{d\times d}$ and $\mathbf{b}_Q,\mathbf{b}_K\in\mathbb{R}^d$ are shared learnable matrix. 
Then we compute the dot product scores $r$ between the metaknowledge representations in query space and key space for each pair of nodes:
\begin{equation}
r_{i,j} = dot(\mathbf{Q}^i, \mathbf{K}^j)
\end{equation}
These scores can then be used to compute the entries of the reconstructed adjacency matrix by using softmax with temperature $\epsilon$ as follows.
\begin{equation}
\mathbf{A}_{i,j} = \frac{exp(r_{i,j}/\epsilon)}{\sum_{k=1}^N exp(r_{i,k}/\epsilon)}
\end{equation}
The reconstructed adjacency matrix encodes the relative importance of the metaknowledge connections between nodes in the traffic graph, with higher values indicating stronger connections.
By using the metaknowledge representations of the nodes, the model is able to capture the more complex and nuanced relationships in the few-shot setting.

\noindent{\textbf{Metaknowledge-auxiliary Forecasting:}}
After reconstructing the adjacency matrix, we can use it to guide the downstream spatial-temporal model (STmodel) in forecasting future traffic.
A normal STmodel takes in near traffic patches $\mathbf{S}_P\in\mathbb{R}^{N\times T_0\times C}$ and the adjacency matrix $\mathbf{A}$ and extracts the traffic information to generate the representations $\mathbf{R}\in\mathbb{R}^{N\times d}$.
The TBP framework could extend to almost every STmodel and we choose Graph Wavenet~\cite{wu2019graph} as our backend due to its simplicity and effectiveness.
Here, the metaknowledge is fused into the adjacency matrix and helps the STmodel capture the dependencies between different nodes.
To further incorporate the metaknowledge into the forecasting process, we concatenate the metaknowledge $\mathbf{M}\in\mathbb{R}^{N\times d}$ with the representation $\mathbf{R}\in\mathbb{R}^{N\times d}$.
The resulting representation is then fed into a regression layer to generate the forecast traffic data $\hat{\mathcal{Y}}$, which is a Multi-Layer Perception (MLP).
This process can be represented as follows.
\begin{equation}
    \begin{aligned}
        \hat{\mathcal{Y}} & =MLP([\mathbf{M}||\mathbf{R}]) \\
                & =MLP([\mathbf{M}||STModel(\mathbf{S}_P,\mathbf{A})])
    \end{aligned}
\end{equation}
Here, $\hat{\mathcal{Y}}\in\mathbb{R}^{N\times T'\times C}$ is the forecasting result. 
Given the ground truth $\mathcal{Y}\in\mathbb{R}^{N\times T'\times C}$, mean square error is selected as the loss:
\begin{equation}
    \label{eq:loss}
    \mathcal{L}=\frac{1}{NT'C}\sum_{i=1}^N\sum_{j=1}^{T'}\sum_{k=1}^{C}(\mathcal{Y}_{ijk}-\hat{\mathcal{Y}}_{ijk})^2
\end{equation}

\noindent{\textbf{Meta-Training \& Fine-tuning:}}
Though the traffic pattern bank contains rich metaknowledge about the traffic patches of source cities, there are other models that need to adapt to the target city such as transformer layer in \textit{Pattern Aggregation}, $\mathbf{Q},\mathbf{K}$ matrices in \textit{Graph Reconstruction}, and STmodel in \textit{Metaknowledge-auxiliary Forecasting}.
These learnable models are collectively referred to as $F_\theta(\cdot)$.
To address this issue, we propose a meta-learning based approach to learn better initial parameters for these models from rich the source city data $G_{source}$.
Then, we fine-tune the model with the few-shot data of the target city $G_{target}$.
We adopt \textit{Reptile}~\cite{nichol2018first} meta-learning framework here, which essentially conducts multi-step gradient descent on the query tasks.
Formally, by denoting the forecasting model as $F_\theta(\cdot)$ and its corresponding parameter as $\theta$, the meta-training and fine-tuning process are shown in Alg.~\ref{alg:meta}.

\section{Experiment}
\definecolor{lgray}{rgb}{0.9,0.9,0.9}

In this section, we evaluate our proposed framework TPB in various aspects through extensive experiments. Specifically, the following research questions are answered. 
\begin{itemize}[leftmargin=*]
    \item \noindent{\textbf{RQ1:}} How does TPB perform against other baselines in the few-shot traffic forecast task?
    \item \noindent{\textbf{RQ2:}} Is TPB able to improve the performance of different STmodels on few-shot forecasting?
    \item \noindent{\textbf{RQ3:}} How to choose the clustering parameter \textit{K} to get a good traffic pattern bank?
    \item \noindent{\textbf{RQ4:}} How does each component of TPB contribute to the final forecasting performance?
\end{itemize}

\subsection{Experiment Settings}
\begin{table}[tp]
    \caption{Statistical details of traffic datasets.}
    \label{tab:data}
    \centering
    \resizebox{\linewidth}{!}{
    \begin{tabular}{c|c|c|c|c}
    \toprule
         & PEMS-BAY & METR-LA & Chengdu & Shenzhen\\
         \midrule
         \# of Nodes & 325 & 207 & 524 & 627\\ 
         \# of Edges & 2,694 & 1,722 & 1,120 & 4,845\\
         Interval & 5 min & 5 min & 10 min & 10 min\\
         \# of Time Step & 52,116 & 34,272 & 17,280 & 17,280\\
         Mean & 61.7768 & 58.2749 & 29.0235 & 31.0092\\
         Std & 9.2852 & 13.1280 & 9.6620 & 10.9694\\ 
    \bottomrule
    \end{tabular}
    }
\end{table}

\noindent{\textbf{Dataset:}}
We evaluate our proposed framework on four commonly used real-world public datasets: \textit{PEMS-BAY}, \textit{METR-LA}~\cite{li2017diffusion}, \textit{Chengdu}, \textit{Shenzhen}~\cite{Didi}.
These datasets contain months of traffic speed data.
The statistical details of these data are listed in Table~\ref{tab:data}.

\noindent{\textbf{Few-shot Setting:}}
We use a similar few-shot traffic forecasting setting to~\cite{lu2022spatio}.
The data of these four cities are divided into source data, target data, and test data.
Here, source data consists of data from three cities, while target and test data consist of data from the target city.
For example, if \textit{PEMS-BAY} is selected as the target city, the full data of \textit{METR-LA}, \textit{Chengdu}, \textit{Shenzhen} constitutes source data.
Then, two-day data of \textit{PEMS-BAY} constitutes the target data and the remaining data of \textit{PEMS-BAY} constitutes the test data.
The Pre-training and Pattern Generation are conducted in the source data, and the Meta-training\&Fine-tuning Process are conducted in the target data.
Finally, we evaluate our framework in the test data.

\noindent{\textbf{Implementation:}}
In Pre-training and Pattern Generation phases, we use $T_0=12$ and $P=24$, which means one-day data is divided into 24 patches to form a patch series.
In the Forecasting stage, we also use the same $T_0$ and $P$ to forecast the future 6 steps of data.
Besides, \textit{Chengdu} and \textit{Shenzhen} have a longer data interval so we conduct linear interpolate to align the positional embedding.
Only speed is contained in the dataset and thus the data channel $C$ is 1.
The mask ratio of Pre-training is set to 75\% according to~\cite{shao2022pre}.
The learning rate of Pre-training is set to 0.0001 and the learning rate of Meta-training $\alpha$ and $\beta$ are both set to 0.0005.
The Adam optimizer of the Fine-tuning has a learning rate of 0.001 and weight decay of 0.01.
The dimension of $H$ and $Key$ and the positional embedding size are set to 128.
The batch size is set to 4.
The number of tasks of meta-training is set to 2.
The experiment is implemented by Pytorch 1.10.0 on RTX3090.
The code is in \url{https://github.com/zhyliu00/TPB}.

\subsection{RQ1: Overall Performance}

\begin{table*}[htb]
\caption{Overall performance of few-shot traffic forecasting on \textit{PEMS-BAY}, \textit{METR-LA}, \textit{Chengdu}, and \textit{Shenzhen}. M,C,S$\rightarrow$PEMS-BAY means the source data is METR-LA, Chengdu, Shenzhen and the target data is PEMS-BAY. The mean and standard deviation of the results in 5 runs is shown. In each column, the best result is highlighted in bold and grey, and the second-best result is underlined. Marker * and ** indicates the mean of the results is statistically significant  (* means t-test with p-value $<$ 0.05 and ** means t-test with p-value $<$ 0.01).}
\label{tab:performance}
\resizebox{0.97\linewidth}{!}{

\begin{tabular}{c|c c c c c c||c c c c c c}
\toprule
& \multicolumn{6}{c||}{\textbf{M,C,S$\rightarrow$PEMS-BAY}} & \multicolumn{6}{c}{\textbf{P,C,S$\rightarrow$METR-LA}}\\
\cline{2-7}
\cline{8-13}
& \multicolumn{2}{c}{5 min} & \multicolumn{2}{c}{15 min} & \multicolumn{2}{c||}{30 min}& \multicolumn{2}{c}{5 min} & \multicolumn{2}{c}{15 min} & \multicolumn{2}{c}{30 min}\\
\cline{2-7}
\cline{8-13}
 &RMSE & MAE & RMSE & MAE & RMSE  & MAE & RMSE & MAE & RMSE & MAE & RMSE  & MAE\\ 

\midrule
\midrule
        HA & 5.49 & 2.67 & 6.01 & 2.95 & 6.56 & 3.23 & 7.37 & 3.97 & 8.05 & 4.27 & 8.76 & 4.66 \\ 
        ARIMA & 4.12 & 2.02 & 4.98 & 2.35 & 5.14 & 2.50 & 4.79 & 3.04 & 6.27 & 3.48 & 7.54 & 4.31 \\ 
        \midrule
DCRNN & 2.38 (0.04) & 1.45 (0.02) & 3.41 (0.07) & 1.90 (0.03) & 4.59 (0.10) & 2.40 (0.06) & 4.76 (0.08) & 2.96 (0.05) & 6.11 (0.09) & 3.45 (0.07) & 7.81 (0.12) & 4.30 (0.06)\\ 
GWN & 2.41 (0.18) & 1.46 (0.10) & 3.52 (0.14) & 1.96 (0.06) & 4.67 (0.10) & 2.36 (0.08) & 4.28 (0.11) & 2.68 (0.17) & 5.72 (0.17) & 3.30 (0.15) & 7.39 (0.26) & 4.11 (0.20)\\ 
STFGNN & 2.31 (0.14) & 1.30 (0.04) & 3.42 (0.07) & 1.81 (0.03) & 4.56 (0.05) & 2.34 (0.07) & 4.39 (0.10) & 2.61 (0.07) & 5.76 (0.07) & 3.17 (0.10) & \underline{7.10 (0.08)} & 3.91 (0.11)\\ 
DSTAGNN & 2.19 (0.17) & 1.32 (0.07) & 3.40 (0.15) & 1.89 (0.14) & 4.87 (0.26) & 2.70 (0.33) & 4.35 (0.08) & 2.59 (0.05) & 5.84 (0.16) & 3.38 (0.08) & 7.68 (0.43) & 4.36 (0.34)\\ 
FOGS & 2.33 (0.23) & 1.36 (0.18) & 3.43 (0.21) & 1.92 (0.16) & 4.53 (0.15) & 2.38 (0.09) & 4.34 (0.12) & 2.56 (0.78) & 6.11 (0.14) & 3.36 (0.10) & 7.40 (0.20) & 3.99 (0.16)\\ 
\midrule
STEP & 2.15 (0.04) & 1.38 (0.02) & 3.26 (0.05) & 1.79 (0.04) & \underline{4.33 (0.08)} & 2.32 (0.04) & 4.26 (0.06) & 2.60 (0.07) & \underline{5.63 (0.09)} & 3.26 (0.09) & 7.18 (0.15) & 3.98 (0.12)\\ 
\midrule
AdaRNN & \underline{1.98 (0.04)} & \underline{1.18 (0.03)} & 3.30 (0.05) & 1.75 (0.04) & 4.40 (0.08) & 2.38 (0.03) & 4.41 (0.14) & 2.60 (0.10) & 5.77 (0.15) & 3.18 (0.10) & 7.33 (0.18) & 3.90 (0.17)\\ 
ST-GFSL & 2.01 (0.08) & 1.18 (0.05) & \underline{3.19 (0.07)} & \underline{1.73 (0.07)} & 4.57 (0.10) & \underline{2.22 (0.06)} & \underline{4.23 (0.09)} & \underline{2.43 (0.08)} & 5.72 (0.12) & \underline{3.03 (0.09)} & 7.28 (0.21) & \underline{3.87 (0.15)}\\ 
\midrule
TPB & \cellcolor{lgray}{\textbf{1.88 (0.02)**}} & \cellcolor{lgray}{\textbf{1.07 (0.02)**}} & \cellcolor{lgray}{\textbf{3.13 (0.04)**}} & \cellcolor{lgray}{\textbf{1.57 (0.02)**}} & \cellcolor{lgray}{\textbf{4.27 (0.06)**}} & \cellcolor{lgray}{\textbf{2.06 (0.03)**}} & \cellcolor{lgray}{\textbf{4.13 (0.07)**}} & \cellcolor{lgray}{\textbf{2.39 (0.06)**}} & \cellcolor{lgray}{\textbf{5.55 (0.09)**}} & \cellcolor{lgray}{\textbf{2.90 (0.07)**}} & \cellcolor{lgray}{\textbf{6.91 (0.12)**}} & \cellcolor{lgray}{\textbf{3.69 (0.08)**}}\\ 

\midrule
\midrule
& \multicolumn{6}{c||}{\textbf{P,M,S$\rightarrow$Chengdu}} & \multicolumn{6}{c}{\textbf{P,M,C$\rightarrow$Shenzhen}}\\
\cline{2-7}
\cline{8-13}
& \multicolumn{2}{c}{10 min} & \multicolumn{2}{c}{30 min} & \multicolumn{2}{c||}{60 min}& \multicolumn{2}{c}{10 min} & \multicolumn{2}{c}{30 min} & \multicolumn{2}{c}{60 min}\\
\cline{2-7}
\cline{8-13}
 &RMSE & MAE & RMSE & MAE & RMSE  & MAE & RMSE & MAE & RMSE & MAE & RMSE  & MAE\\ 
\midrule
\midrule
        HA & 4.29 & 3.02 & 4.80 & 3.42 & 5.44 & 3.91 & 3.81 & 2.55 & 4.21 & 2.82 & 4.73 & 3.19 \\ 
        ARIMA & 3.79 & 2.75 & 4.40 & 3.21 & 5.09 & 3.60 & 3.38 & 2.34 & 4.10 & 2.78 & 4.68 & 3.08 \\ 
        \midrule
DCRNN & 3.27 (0.07) & 2.28 (0.06) & 4.13 (0.07) & 2.89 (0.07) & 4.78 (0.10) & 3.39 (0.08) & 2.81 (0.05) & 1.94 (0.04) & 3.59 (0.07) & 2.41 (0.06) & 4.11 (0.08) & 2.78 (0.05)\\ 
GWN & 3.33 (0.15) & 2.34 (0.14) & 4.22 (0.09) & 2.94 (0.09) & 4.87 (0.14) & 3.50 (0.12) & 2.88 (0.06) & \underline{1.93 (0.05)} & 3.68 (0.08) & 2.50 (0.07) & 4.36 (0.08) & 2.86 (0.06)\\ 
STFGNN & 3.33 (0.08) & 2.33 (0.06) & 4.19 (0.06) & 2.96 (0.06) & 4.90 (0.12) & 3.47 (0.07) & 2.89 (0.04) & 1.97 (0.07) & 3.67 (0.07) & 2.50 (0.09) & 4.27 (0.07) & 2.89 (0.08)\\ 
DSTAGNN & 3.31 (0.19) & 2.30 (0.14) & 4.15 (0.12) & 2.99 (0.11) & 4.90 (0.16) & 3.47 (0.10) & 3.00 (0.08) & 1.99 (0.05) & 3.65 (0.12) & 2.48 (0.10) & 4.32 (0.09) & 2.91 (0.07)\\ 
FOGS & 3.26 (0.09) & 2.27 (0.09) & 4.21 (0.12) & 2.88 (0.11) & 4.71 (0.11) & 3.29 (0.09) & 2.85 (0.08) & 1.96 (0.07) & 3.68 (0.10) & 2.38 (0.08) & 4.28 (0.12) & 2.86 (0.10)\\ 
\midrule
STEP & \underline{3.19 (0.05)} & \underline{2.26 (0.04)} & \underline{3.91 (0.08)} & \underline{2.75 (0.07)} & \underline{4.31 (0.11)} & \underline{3.15 (0.10)} & \underline{2.80 (0.05)} & 1.98 (0.04) & \underline{3.47 (0.05)} & \underline{2.38 (0.06)} & \cellcolor{lgray}{\textbf{3.77 (0.07)}} & \underline{2.52 (0.06)}\\ 
\midrule
AdaRNN & 3.24 (0.08) & 2.26 (0.07) & 4.05 (0.08) & 2.95 (0.08) & 4.90 (0.08) & 3.50 (0.07) & 2.86 (0.06) & 1.98 (0.06) & 3.57 (0.07) & 2.44 (0.06) & 4.20 (0.08) & 2.84 (0.07)\\ 
ST-GFSL & 3.37 (0.12) & 2.29 (0.09) & 4.20 (0.10) & 2.86 (0.09) & 4.68 (0.12) & 3.28 (0.10) & 2.97 (0.09) & 2.00 (0.08) & 3.73 (0.07) & 2.40 (0.07) & 4.25 (0.10) & 2.76 (0.07)\\ 
\midrule
TPB & \cellcolor{lgray}{\textbf{3.05 (0.04)**}} & \cellcolor{lgray}{\textbf{2.10 (0.04)**}} & \cellcolor{lgray}{\textbf{3.80 (0.06)**}} & \cellcolor{lgray}{\textbf{2.65 (0.04)**}} & \cellcolor{lgray}{\textbf{4.30 (0.07)**}} & \cellcolor{lgray}{\textbf{3.02 (0.06)**}} & \cellcolor{lgray}{\textbf{2.68 (0.04)**}} & \cellcolor{lgray}{\textbf{1.80 (0.03)**}} & \cellcolor{lgray}{\textbf{3.32 (0.05)**}} & \cellcolor{lgray}{\textbf{2.22 (0.07)**}} & \underline{3.80 (0.06)**} & \cellcolor{lgray}{\textbf{2.50 (0.07)**}}\\ 

\bottomrule
\bottomrule

\end{tabular}
}
\end{table*}
\noindent{\textbf{Baselines:}}
We select 10 baselines of four types to evaluate the performance of TPB on the few-shot forecasting task. 
These baselines include traditional methods, typical deep-learning traffic forecasting methods, pre-training traffic forecasting methods, and cross-city traffic forecasting methods.
We use the model structure hyper-parameters reported by the original papers of these models or frameworks. 
Note that the typical deep-learning traffic forecasting methods are implemented in \textit{Reptile}, which is a meta-learning framework. 
We guarantee the fairness of the comparisons of different baselines.
For the hyper-parameter of meta-learning framework, we hold the same sufficient strategy for meta-learning hyperparameters searching in all methods including TPB to guarantee fairness.
The learning rate of meta-training \& fine-tuning is set to 5e-4, the meta-training epoch is searched in range(5, 30, 5), the fine-tuning epoch is searched in range(50, 400, 50), update\_step is searched in range(2, 5, 1) (larger update\_step would lead to GPU memory limit exceeded).


\begin{itemize}[leftmargin=*]
    \item Traditional methods
    \begin{itemize}[leftmargin=*]
        \item \textbf{HA}: Historical Average, which uses the average of previous periods as the predictions of future traffic.
        \item \textbf{ARIMA}~\cite{williams2003modeling}: A commonly used statistical time series forecasting model.
    \end{itemize}
    \item Typical deep-learning traffic forecasting methods (\textit{Reptile})
    \begin{itemize}[leftmargin=*]
        \item \textbf{DCRNN}~\cite{li2017diffusion}: A spatial-temporal network that uses diffusion techniques and RNN.
        \item \textbf{GWN}~\cite{wu2019graph}: Graph WaveNet utilizes the adaptive adjacency matrix and dilated causal convolution to capture the spatial temporal dependency.
        \item \textbf{STFGNN}~\cite{li2021spatial}: A network that uses DTW distance to construct a temporal graph and fuse the spatial and temporal information in a gated module.
        \item \textbf{DSTAGNN}~\cite{lan2022dstagnn}: A network that constructs spatial-temporal aware graph and uses K-th order Chebyshev polynomial to aggregate the information.
        \item \textbf{FOGS}~\cite{rao2022fogs}: A framework that uses node2vec to learn spatial-temporal correlation and predicts the first-order gradients.
    \end{itemize}
    \item Pre-training traffic forecasting method
    \begin{itemize}[leftmargin=*]
        \item \textbf{STEP}~\cite{shao2022pre}: A pre-training framework that reconstructs the large-ratio masked traffic data based on transformer layers.
    \end{itemize}
    \item Cross-city traffic forecasting methods
    \begin{itemize}[leftmargin=*]
        \item \textbf{AdaRNN}~\cite{du2021adarnn}: A time series transfer learning framework that reduces the distribution mismatch between time series to make model more adaptive.
        \item \textbf{ST-GFSL}~\cite{lu2022spatio}: A state-of-art few-shot traffic forecasting framework that learns the metaknowledge of traffic nodes to generate the parameter of linear and convolution layers.
    \end{itemize}
\end{itemize}

Not that baselines such as STrans-GAN~\cite{zhang2022strans}, CrossTReS~\cite{jin2022selective} could not be fairly compared here since they utilize multimodal data while our setting is cross-city univariate forecasting. The reason to do univariate forecasting is that we would like to tackle the situation that the target city has very few univariate data and data from other modalities is missing.

\noindent{\textbf{Metric:}}
Two commonly used metrics in traffic forecasting are used to evaluate the performance of all baselines, including Root Mean Squared Error (RMSE) and Mean Absolute Error (MAE). The formulas are as follows.
$$
    RMSE=\sqrt{\frac{1}{s}\sum_{i=1}^s(y_i-\hat{y_i})^2},\ \ MAE=\frac{1}{s}\sum_{i=1}^s|{y_i-\hat{y_i}}|
$$

\noindent{\textbf{Results:}}
The overall result is shown in Table~\ref{tab:performance}. 
The following observations could be drawn:
(1) Our proposed TPB achieves superior performance compared to the baselines not only in short-term but also in long-term forecasting.
(2) Traditional methods have bad performance because the forecasting task has strong non-linearity, which is hard to be captured by these methods.
(3) Compared with the typical deep-learning traffic forecasting methods implemented in the \textit{Reptile} framework, TPB achieves a performance enhancement of 9.92\% and 13.13\% in RMSE and MAE respectively on average. 
The difference between these methods and our framework is that we add traffic-pattern-based metaknowledge in the forecasting stage.
So, the significant performance enhancement indicates that the additional metaknowledge offered by the traffic pattern bank does help the model forecast more precisely.
Further experiments about the effectiveness of our proposed traffic pattern bank would be shown in Sec.~\ref{addtpb}
(4) Compared with the pre-training traffic forecasting method, TPB achieves a performance enhancement of 4.26\% and 8.96\% in RMSE and MAE respectively.
This indicates that directly using the pre-trained transformer encoder is not enough and refining the metaknowledge in encoder space to construct the traffic pattern bank does help the model make accurate predictions in the few-shot traffic forecasting task.
(5) Compared with the cross-city traffic forecasting methods, TPB achieves a performance enhancement of 6.52\% and 7.71\% in RMSE and MAE respectively on average and outperforms these methods on all datasets.
The result shows that the traffic pattern bank of the TPB framework captures better metaknowledge than the unique design of ST-GFSL and AdaRNN.

\begin{figure}[tp]
    \centering
    \includegraphics[width=0.94\linewidth]{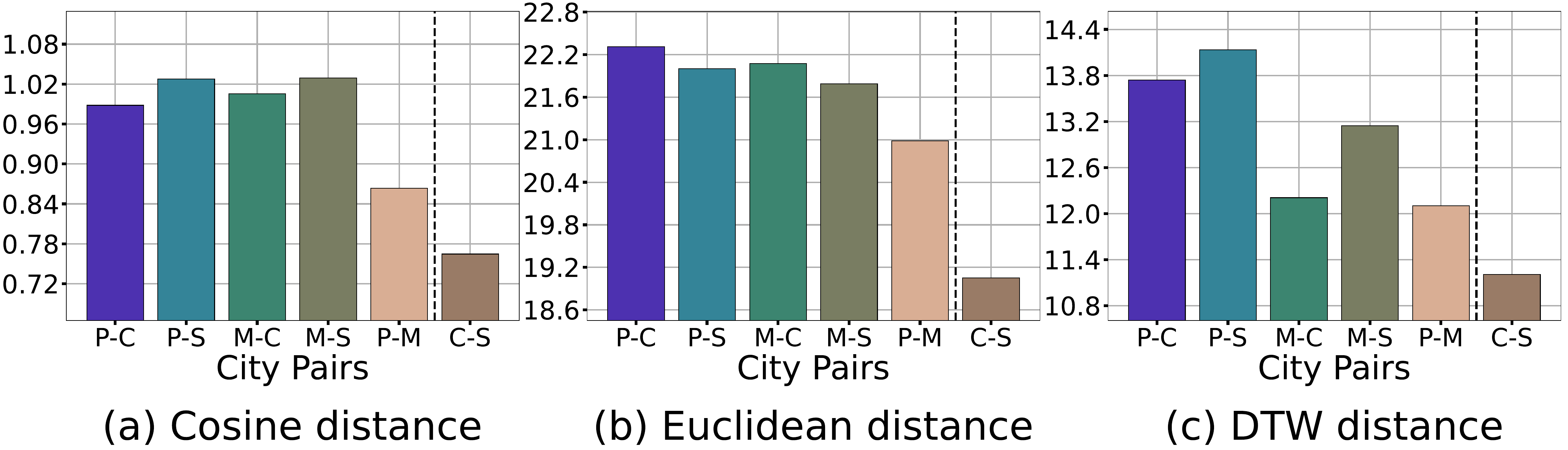}
    \caption{The distance between each pair of one-day raw traffic patches (time series of length 288) of four datasets. For simplicity, datasets are denoted by their first letter, e.g., "C-S" indicates \textit{Chengdu} and \textit{Shenzhen}. 
    }
    \label{fig:citypair}
\end{figure}

\noindent{\textbf{Data Distribution Analysis:}} Fig.~\ref{fig:citypair} shows the dataset distance between \textit{Chengdu} and \textit{Shenzhen} is smaller than other dataset pairs on all of the three metrics and Table~\ref{tab:data} shows the mean and std between \textit{Chengdu} and \textit{Shenzhen} are more similar.
So, The data distributions of \textit{Chengdu} and \textit{Shenzhen} are more similar than other pairs of cities.
Consequently, the setting of P,M,S$\rightarrow$Chengdu and P,M,C$\rightarrow$Shenzhen represent the setting that knowledge transfers to target city with similar distribution while M,C,S$\rightarrow$PEMS-BAY and P,C,S$\rightarrow$METR-LA represent the setting that knowledge transfers to a target city with different distribution.
We could observe that TPB outperforms other methods in both settings in Table~\ref{tab:performance}.
This demonstrates the superiority and robustness of TPB in the cross-city forecasting scenario.

\subsection{RQ2: TPB Improves Different STmodel}
\label{addtpb}

TPB is a framework for few-shot traffic forecasting that different STmodels are able to plug into this framework.
Specifically, TPB offers the reconstructed adjacency matrix $A$ and metaknowledge embedding $M$ based on traffic pattern bank to help the STmodel forecasting precisely.
In order to verify the effectiveness of the traffic pattern bank on the few-shot setting, we implement several advanced STmodels into the TPB framework and check the performance enhancement.
We have two types of training processes for these STmodels.
(1) These STmodels are trained in \textit{Reptile} meta-learning framework and the results are denoted as \textit{STmodel-Reptile}.
(2) These STmodels are plugged into the TPB framework and the training process of TPB is then conducted, the results of which are denoted as \textit{STmodel-TPB}.
By comparing the results of these two types of setting, we could know whether the traffic pattern bank help forecast future traffic data.

\begin{figure}[tp]
    \centering
    \includegraphics[width=0.94\linewidth]{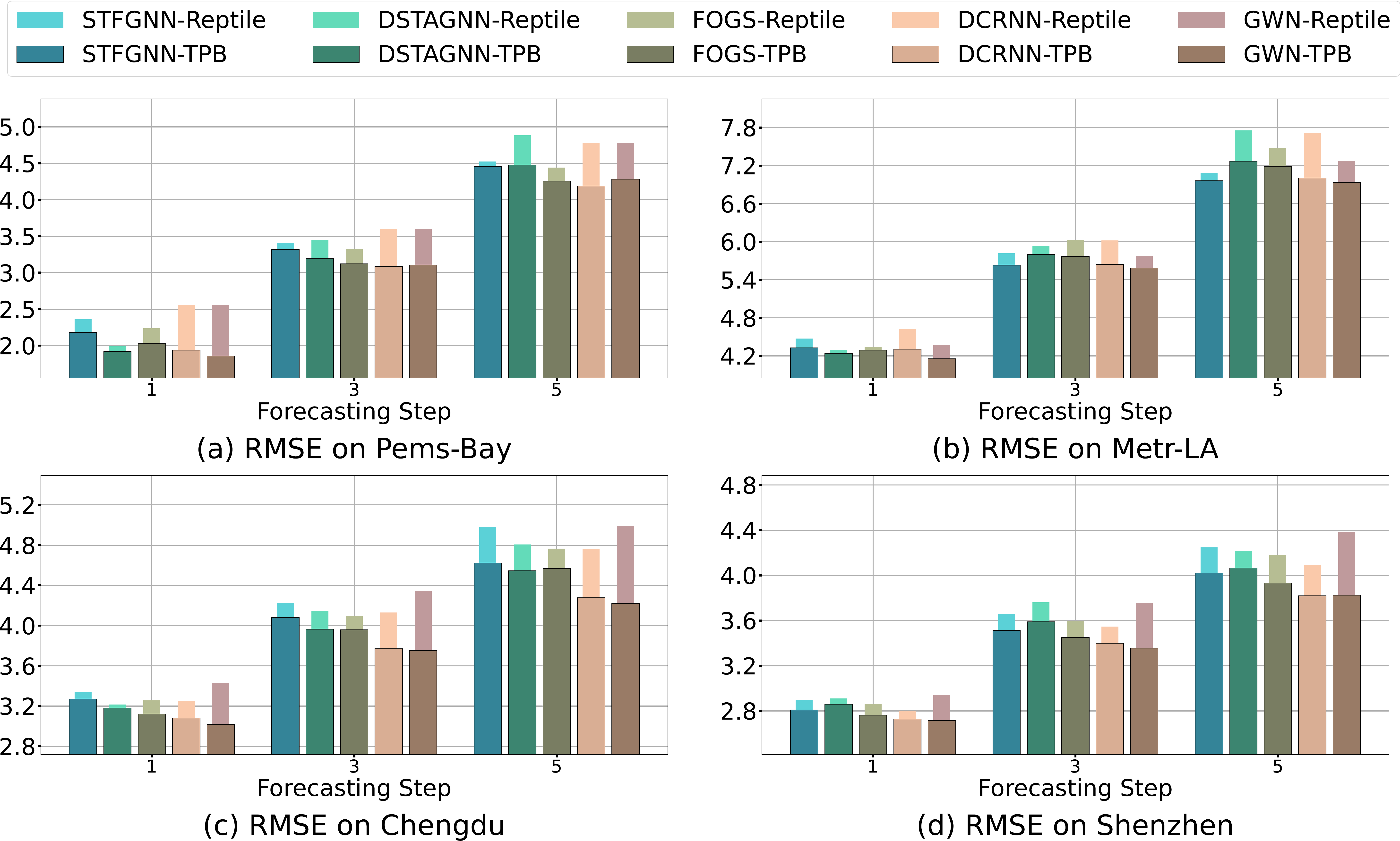}
    \caption{Performance comparison of different STmodels between being trained in \textit{Reptile} meta-learning framework and being trained in TPB framework. RMSE of multi-step forecasting on four datasets is plotted.}
    \label{fig:addTPB}
\end{figure}

Fig.~\ref{fig:addTPB} shows the RMSE results of the aforementioned methods on four datasets of multi-step forecasting, from which we have the following observations:
(1) Plugged into the TPB framework, all of the STmodels show improved performance compared to when they are trained in the \textit{Reptile} meta-learning framework. 
This demonstrates the effectiveness of the traffic pattern bank in helping the STmodels to perform better on the few-shot traffic forecasting task.
(2) The performance enhancement is more pronounced for longer forecasting horizons.
This indicates the traffic pattern bank is able to capture more complex patterns that are important for longer-term forecasting.
(3) Simple STmodels such as Graph Wavenet have shown more significant performance improvement than complex STmodels such as STFGNN and DSTAGNN. 
This indicates that simple STmodels are more flexible and the metaknowledge of traffic pattern bank adapts to these models more easily.

\subsection{RQ3: How to choose \textit{K}?}

\begin{figure*}[thp]
    \centering
    \includegraphics[width=0.9\linewidth]{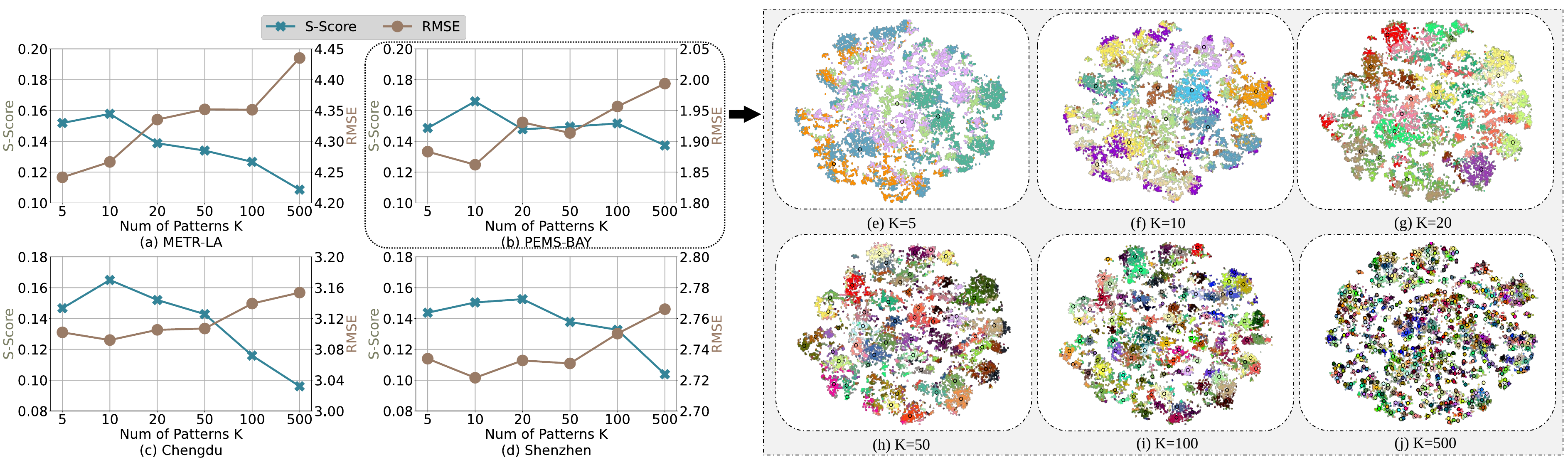}
    \caption{Patterns analysis. (a)$\sim$(d) The Silhouette score of clustering result and RMSE of the 1-hop forecasting results of different clustering parameters \textit{K} on four datasets. A higher Silhouette score indicates the clusters are better defined and thus the traffic pattern bank is of better quality. (e)$\sim$(j) The TSNE visualization of the embeddings of traffic patch and pattern with different \textit{K} on the PEMS-BAY dataset. The colored dots indicate the traffic patch embeddings belonging to different clusters. The black circles indicate the traffic pattern embeddings, which are the centroids of different clusters.}
    \label{fig:pattern}
\end{figure*}

In the pattern generation phase, clustering parameter \textit{K} significantly influences the quality of the generated traffic pattern bank.
To quantitatively analyze the quality of the traffic pattern bank of different \textit{K}, we calculate the Silhouette score~\cite{rousseeuw1987silhouettes} of the traffic pattern bank. It is calculated as follows.
\begin{equation}
  s = \sum_{i=1}^{P}\frac{b_i - a_i}{\max(a_i, b_i)}  
\end{equation}
Here, $P$ is the total number of traffic patches and $a_i$ is the average distance between the traffic patch and all other traffic patches in its cluster and $b_i$ is the average distance between the traffic patch and all other traffic patches in the nearest cluster. 
The Silhouette score ranges from -1 to 1, where a higher value indicates that the samples within each cluster are very similar to one another and thus the clusters are well-defined.

Fig.~\ref{fig:pattern} shows two parts of the result. 
In (a)$\sim$(d), the Silhouette score of clustering results and RMSE of the 1-hop forecasting results of different \textit{K} on four datasets are plotted. 
The following phenomena could be observed in these four plots.
(1) When \textit{K} is small (around 10), the Silhouette scores of the clustering result are large, and the RMSE of these results is small.
(2) As \textit{K} increases, the Silhouette score drops drastically, and the RMSE increases drastically at the same time.
From these phenomena, we can say that the Silhouette score is a good criterion for the quality of the traffic pattern bank given a \textit{K}, and using a traffic pattern bank with a high Silhouette score in TBP could lead to a small error.
In (e)$\sim$(i), the TSNE visualization of the embeddings of traffic patch and pattern with different \textit{K} on the PEMS-BAY dataset is shown.
From these plots, we could observe that the traffic patch embeddings belonging to different clusters are more distinct when a smaller value of \textit{K} is used, while a larger value of \textit{K} results in one raw cluster being divided into several traffic patterns. 
This indicates that a smaller value of \textit{K} produces more well-defined clusters and a higher-quality traffic pattern bank.
Furthermore, we could observe that the final result of TPB is inversely proportional to the Silhouette score.
Consequently, it is efficient to select a proper \textit{K} by calculating the Silhouette score and choosing the \textit{K} with the highest Silhouette score.



\subsection{RQ4: Ablation Study}

\begin{figure}[tp]
    \centering
    \includegraphics[width=0.98\linewidth]{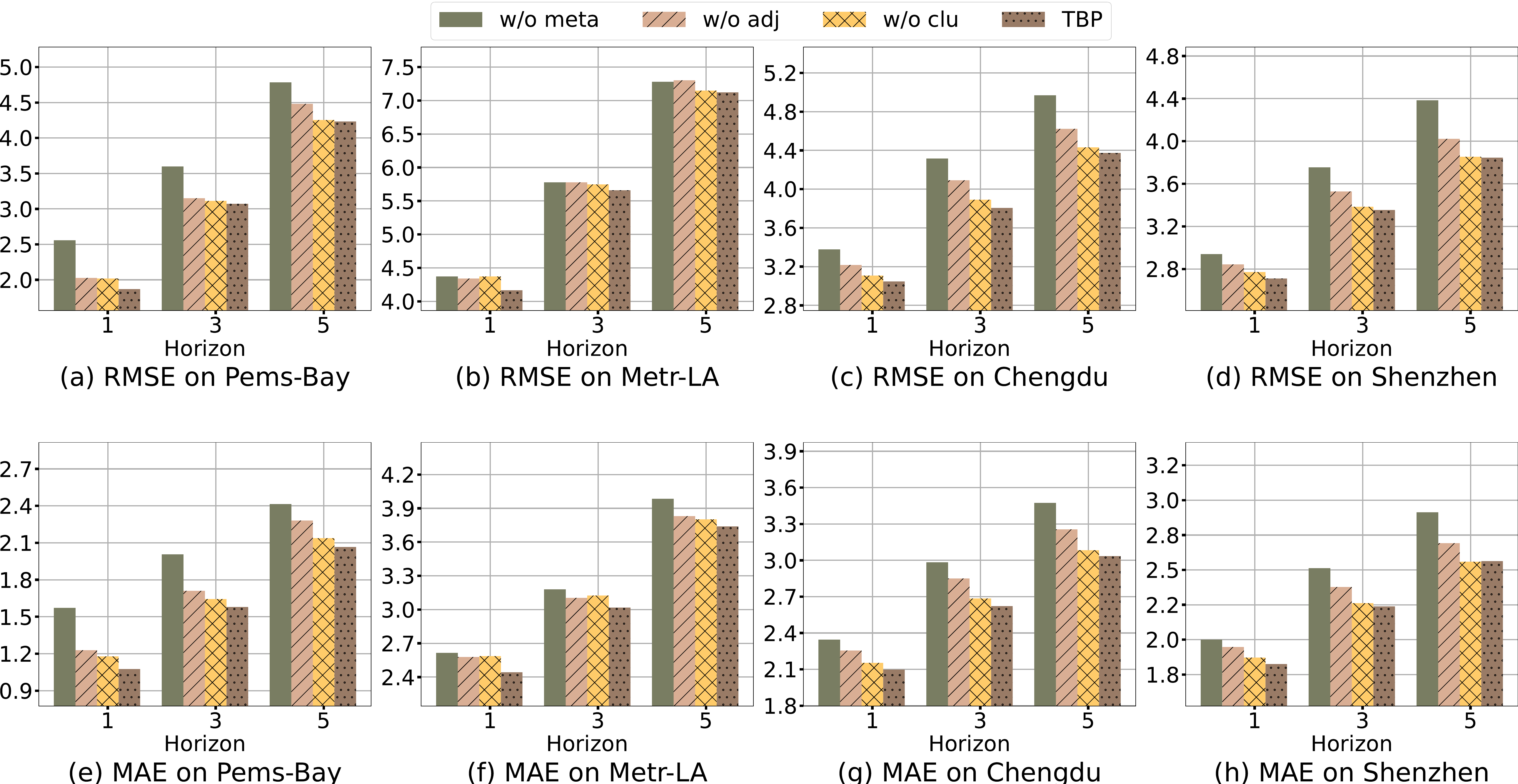}
    \caption{Ablation study of TPB. Both RMSE and MAE of multi-step forecasting of the variants are reported.}
    \label{fig:ablation}
    \vspace{-0.2cm}
\end{figure}

In this section, we verify the effectiveness of each module of TPB.
We remove or modify each module to get a variant of TPB and do grid search to get the best performance of these variants.
There are three types of variants of TPB.
(1) We remove traffic pattern bank and downstream metaknowledge. This variant is denoted as \textbf{w/o meta}.
(2) We replace the metaknowledge-based adjacency matrix with the pre-defined static adjacency matrix, which is denoted as \textbf{w/o adj}.
(3) We remove clustering process and randomly select \textit{K} traffic patches as traffic pattern bank, which is denoted as \textbf{w/o clu}.

Fig.~\ref{fig:ablation} shows the results of these three variants and compares them with the results of TPB.
First, we can see that without the metaknowledge, the performance of the variant has significantly decreased, which demonstrates traffic pattern bank contains rich metaknowledge from the source cities and is helpful in few-shot traffic forecasting.
Then, the variant that uses the predefined graph performs worse.
This is not surprising because the predefined graph is not able to capture the dynamic and context-specific relationships between nodes, while the metaknowledge-based adjacency matrix is constructed based on the traffic pattern bank and is able to adapt to the specific spatial relation of the target city.
Finally, the random-selected traffic pattern bank degrades the performance.
This indicates the clustering process selects robust and representative traffic patterns from the source cities and the use of these patterns in the target city leads to improved performance.



\section{Conclusion}
In this work, we propose a novel cross-city few-shot traffic forecasting framework named TPB.
We demonstrate that the traffic pattern is similar across cities.
To capture the similarity, we pre-train a traffic patch encoder and use it to generate a traffic pattern bank from data-rich cities to help downstream cross-city traffic forecasting.
Experiments on real-world traffic datasets demonstrate the superiority over state-of-the-art methods for cross-city few-shot traffic forecasting of the TPB framework.
In the future, we will further investigate the usage of pattern bank on other few-shot settings.

\section*{Acknowledgement}

This work was sponsored by National Key Research and Development Program of China under Grant No.2022YFB3904204, National Natural Science Foundation of China under Grant No.62102246, No.62272301, No.62176243, and Provincial Key Research and Development Program of Zhejiang under Grant No.2021C01034.

\bibliographystyle{named}
\clearpage
\balance
\bibliography{cikm22}

\end{document}